\documentclass[letterpaper]{article} 
\usepackage{aaai25}  
\usepackage{times}  
\usepackage{helvet}  
\usepackage{courier}  
\usepackage[hyphens]{url}  
\usepackage{graphicx} 
\usepackage{natbib}  
\usepackage{caption} 
\usepackage{algorithm}
\usepackage{algorithmic}

\usepackage{amsmath}
\usepackage{amsfonts}
\usepackage{fdsymbol}
\usepackage{makecell}
\usepackage{multirow}
\usepackage{booktabs}
\usepackage{bm}

\usepackage{newfloat}
\usepackage{listings}
\DeclareCaptionStyle{ruled}{labelfont=normalfont,labelsep=colon,strut=off} 
\floatstyle{ruled}
\newfloat{listing}{tb}{lst}{}
\floatname{listing}{Listing}
\title{Uncertainty-Aware Global-View Reconstruction for \\ Multi-View Multi-Label Feature Selection}
\author{
    Pingting Hao\textsuperscript{\rm 1,2}
    Kunpeng Liu\textsuperscript{\rm 3},
    Wanfu Gao\textsuperscript{\rm 1,2}\thanks{correponding author}
}
\affiliations{
    \textsuperscript{\rm 1} College of Computer Science and Technology, Jilin University, China\\

   \textsuperscript{\rm 2} Key Laboratory of Symbolic Computation and Knowledge Engineering of Ministry of
Education, Jilin University, China\\
 \textsuperscript{\rm 3} Department of Computer Science, Portland State University, Portland, OR 97201 USA\\
    haopingting@jlu.edu.cn, kunpeng@pdx.edu, gaowf@jlu.edu.cn
}

\usepackage{bibentry}

\begin{document}

\maketitle

\begin{abstract}
In recent years, multi-view multi-label learning (MVML) has gained popularity due to its close resemblance to real-world scenarios. However, the challenge of selecting informative features to ensure both performance and efficiency remains a significant question in MVML. Existing methods often extract information separately from the consistency part and the complementary part, which may result in noise due to unclear segmentation. In this paper, we propose a unified model constructed from the perspective of global-view reconstruction. Additionally, while feature selection methods can discern the importance of features, they typically overlook the uncertainty of samples, which is prevalent in realistic scenarios. To address this, we incorporate the perception of sample uncertainty during the reconstruction process to enhance trustworthiness. Thus, the global-view is reconstructed through the graph structure between samples, sample confidence, and the view relationship. The accurate mapping is established between the reconstructed view and the label matrix. Experimental results demonstrate the superior performance of our method on multi-view datasets.
\end{abstract}

%

\section{Introduction}
Feature selection plays a vital role in various domains, including healthcare ~\cite{bharati2023review}, anomaly detection ~\cite{zhang2024realnet}, and transportation ~\cite{wang2023segmentalized}. It has been applied in different scenarios, such as unsupervised ~\cite{zhang2024efficient}, single-label ~\cite{cohen2023few}, and multi-label ~\cite{klonecki2023cost} scenarios, based on specific requirements. The challenge of selecting the optimal subset of features is particularly daunting in the presence of high-dimensional data. However, the multi-view setting offers the advantage of rich semantics combined with more comprehensive data, which in turn introduces additional complexities for feature selection in MVML.

\begin{figure*}[t]
\centering
\includegraphics[width=1.9\columnwidth]{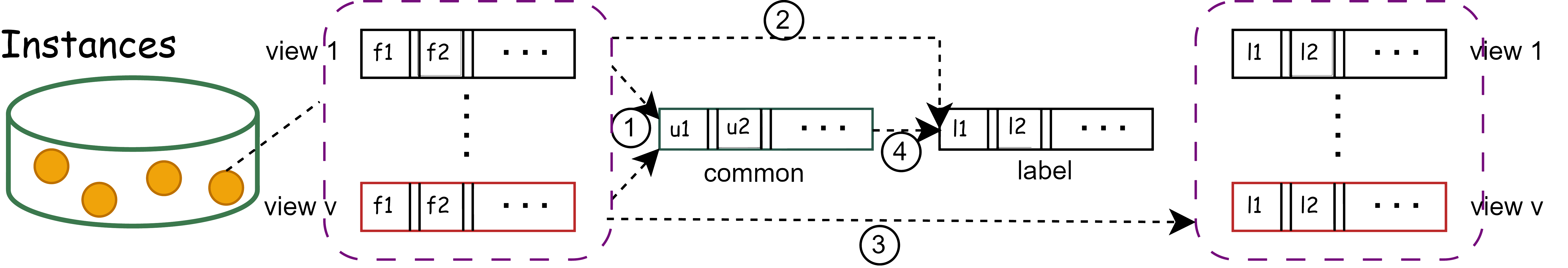} 
\caption{Four typical types for the relationship between features and labels in MVML, namely, a) common-based mapping including $\textcircled{1}$ and $\textcircled{4}$, b) label consistency mapping including $\textcircled{2}$, c) view-specific mapping including $\textcircled{3}$, and d) concatenating view mapping including $\textcircled{4}$.}
\label{fig1}
\end{figure*}

Differentiating from the single-view multi-label scenario, effectively managing the view relationship and integrating information from multiple views is a fundamental aspect in multi-view scenario ~\cite{lyucommon1, lyu2024vsm}. Previous works in the multi-view scenario primarily focus on identifying commonality across views as an initial step, in accordance with the defined consistency rules between views ~\cite{xu2013survey,sun2013survey}. Typically, the consensus part is derived from the features ~\cite{zhang2018latent,liu2023label,yin2023multi,liu2015low}, as depicted in Figure 1a. However, the distinctive part is often overlooked, resulting in a deficiency of information and subsequent performance degradation of the model.

Given the increasing recognition of the significance of the distinctive part within each view, it is imperative to acknowledge the exploration of individual information. Noteworthy works, such as ~\cite{liu2023multi,li2021concise}, operate under the assumption of label matrix consistency for each view and establish mappings between each view and the label matrix, as depicted in Figure 1b. Alternatively, Tan et al. ~\cite{tan2019individuality} construct separate mappings for common and individual views, thereby addressing both aspects. In reality, the hypothesis of an inconsistent label matrix based on each view, as illustrated in Figure 1c, can also be considered. Zhao et al. ~\cite{zhao2022learning} partition the original label matrix into multiple matrices corresponding to each view to preserve the distinctive information within each view. Nonetheless, the accuracy of the process for solving feature weights is compromised due to biased grounds or incomplete constraints. This is primarily because the perspective is not grounded on the ground-truth of the entire view to avoid unclear segmentation.

Although some works such as ~\cite{zhu2020global} assume that the label matrix for the whole dataset and each view are observed, and subsequently establish mappings between features and labels for both. However, they often overlook the view relationship and fail to consider the sample-level relationship, as depicted in Figure 1d. Thus, there most likely exists noise and redundancy in direct concatenating matrix. How to utilize effective information to reconstruct the whole dataset becomes challenging.

Indeed, it is important to acknowledge that the label information inherently encompasses valuable comprehensive information. This distinguishes it from the unsupervised multi-view scenario. The effective utilization of this information significantly impacts the performance of feature selection models. Recent works, such as ~\cite{zhu2015block} and ~\cite{zhang2020multi}, leverage the inherent integrity property of the label matrix to establish a mapping relationship between multiple views and a single label matrix. Nevertheless, the effective learning of weights for all features remains a challenging task due to the inherent many-to-one mapping relationship.

Moreover, it is worth considering the relaxation of the assumption that the reliability of each view, and even each sample, is uniformly equal to 1. In reality, the quality of histopathological images may vary across different patients ~\cite{yagi2011color}, and tabular data may contain feature noise ~\cite{yelipe2018efficient}. As a result, there can be disparate confidence levels among samples. Therefore, the ability to perceive the informativeness of different samples can enhance the explainability ~\cite{han2022multimodal}, and facilitate the identification of correct samples for potential data adjustment and retraining. This is particularly crucial in multi-view multi-label feature selection as it helps eliminate interference caused by samples, and mitigates the introduction of noise when establishing the mapping for feature weights.

To tackle the above issues, we propose a novel method called \textbf{U}ncertainty-aware \textbf{G}lobal-view \textbf{R}econstruction for Multi-view Multi-label \textbf{F}eature \textbf{S}election (UGRFS). By taking into account the complete information contained in the label matrix, we introduce a regularizer paradigm that captures the relationship between the higher-dimensional label space and the initialization of the fusion feature space using view weights. In addition, our method incorporates constraints that operate on the graph structure of the global-view and its splitting structure with uncertainty. Overall, the main contributions of this paper can be summarized as follows:
\begin{itemize}
    \item The UGRFS method introduces a sparse model that perceives sample confidence in the multi-view multi-label scenario. Its objective is to tackle noise and redundancy across views. This finer-grained perception not only facilitates evaluation and adjustment based on expert knowledge, but also promotes with the computation of feature weights mutually to improve performance.
    \item The regularization paradigms for global-view reconstruction aim to establish an accurate mapping that considers the graph structure, sample-relationship, and view-relationship. In our method, the new variable for global-view is automatically learned instead of being fused with predefined weights. This allows for balancing the consistency and complementary within one variable.
    \item To ensure algorithm convergence, the multiplicative update rules are applied to the optimization process. Subsequently, extensive experiments are conducted on six multi-view multi-label datasets. The experimental results validate the superiority of UGRFS over state-of-the-art methods.
\end{itemize}
\section{Related Work}

\subsection{Multi-Label Feature Selection}
In the multi-label scenario, instances are associated with multiple labels to capture their rich semantics ~\cite{zhangjia2020multi}. Unlike single-label or unsupervised methods, some existing multi-label feature selection methods focus on the label matrix. Specific to the noises in label matrix, Jian et al. ~\cite{jian2016multi} utilize low-rank techniques to obtain the latent label matrix. Additionally, the reconstruction of the label matrix utilizes label correlation and the dependence of the original label space to constrain its graph structure, as demonstrated in previous works ~\cite{fan2021manifold,huang2021multi,han2022multimodal}.

Considering mainly for feature selection problem, how to deal with the feature weights is of primary importance for multi-label scenario. The design of the regularizer for feature weights considers structure constraints and leverages the benefits of different norms. Some methods incorporate label correlation as a constraint or combine multiple paradigms ~\cite{huang2019improving,li2022learning,li2023multi}. Alternatively, other techniques decompose the feature weight matrix to enhance the relationship between features and labels ~\cite{lin2023multi}, or extract local discriminative features ~\cite{fan2021multi}. Although the multi-label problem shares similarities with MVML, the distinction lies in the incorporation of multiple views.

\subsection{Multi-View Multi-Label Learning}
Expansion based on multi-label scenario, the cases for diverse description of data has explosive growth ~\cite{sanghavi2022multi}. Other expansion cases are also discussed based on MVML such as the non-aligned case ~\cite{zhong2024align} and incomplete views.  Combined with label matrix, complex cases with incomplete views include but not limited to missing labels ~\cite{liu2023dicnet}, label noises ~\cite{liu2022incomplete} and partial labels ~\cite{jiang2024multiview}. Existing MVML methods strive to balance consistency and complementarity to preserve the comprehensive information ~\cite{lyu2022beyond,wu2019multi}. Zhu et al. ~\cite{zhu2018multi} propose a method to incorporate diverse information from the feature space and label space into the latent space of each view, deviating from traditional loss functions in MVML.
\section{The Proposed Method}

\subsection{Problem Formulation}
Given a multi-view multi-label dataset $(X, Y)$, $X=\left \{ {{X^{\left(i\right)}}} \right \} _{i=1}^V$ contains $V$ different views, and $X^{(i)}\in\mathbb{R}^{n\times d(i)}$ with the same number of instances $n$ in different views. Each element $x_{m.}^{(i)}\in X^{(i)}$ is a $d(i)$-dimensional feature vector and denotes the $m$-th instance for $i$-th view. The fusion view with view-relationship is setting as $X^f\in\mathbb{R}^{n\times d}$, and the global-view distribution represents as $D\in\mathbb{R}^{n\times d}$ where $d$ denotes the number of features. The sample confidence matrix is defined as non-negative variable $C=\left \{ {{C^{\left(i\right)}}} \right \} _{i=1}^V$ and $C^{(i)}\in\mathbb{R}^{n\times 1}$. In the observed label matrix $Y\in\left\{0,1\right\}^{n\times l}$, the $j$-th label is denoted by $y_{.j}$. In this section, the main procedure of UGRFS is introduced, including global-view reconstruction $L_G$ and uncertainty-aware feature selection $L_U$, which can be formalized: $L_{URGFS}=L_G(\cdot)+L_U(\cdot)$.

\subsection{Global-View Reconstruction}
\subsubsection{View-Relationship Fusion}
Due to the uneven distribution of information content among different views, directly concatenating them easily ignores the difference between views. Instead, estimating the contribution of each view to the global-view matrix can be achieved through the computation of similarity in local geometric structures.

To compute the similarity of the local structure, we exploit it in the feature space. The affinity matrix, denoted as ${S}^{({v})}$, is computed for the features in the $v$-th view as a core component of graph Laplacian ~\cite{chung1997spectral}. The element in the matrix is defined as:
\begin{gather}
s_{ij}^{(v)}=\left\{
\begin{array}{rcl}
exp(-\frac{{||x_{i.}^{(v)}-x_{j.}^{(v)}||}_2^2}{\sigma^2})&{if\ x_{i.}^{(v)}\in\mathcal{N}_q({x}_{j.}^{(v)})\ or } \\
\quad  & {{x}_{j.}^{(v)}\in\mathcal{N}_q(x_{i.}^{(v)})} \\
0  & {otherwise}
\end{array} \right.,
\end{gather}
where $\sigma$ represents a predefined parameter, and $\mathcal{N}_q(x_{j.}^{(v)})$ represents the set of $q$ nearest neighbors of $x_{i.}^{(v)}$.  Similarly, the meaning of $\mathcal{N}_q(x_{i.}^{(v)})$ can be inferred. The formula based on the smooth assumption ~\cite{zhu2005semi} can be represented as:
\begin{gather}\label{3}
  \frac{1}{2}\sum_{i=1}^{n}\sum_{j=1}^{n}{s_{ij}^{(v)}\left(y_{i.}-y_{j.}\right)^2}.
\end{gather}

According to the definition of the graph Laplacian, the graph Laplacian matrix ${L}_{x}^{({v})}={A}^{({v})}-{S}^{({v})}$, where ${A}^{({v})}$ is the diagonal matrix, whose elements are $a_{ii}^{({v})}=\sum_{j=1}^{n}s_{ij}^{({v})}$. Thus, Formula (2) can be transformed into Formula (3) as:
\begin{gather}
Tr(y^T\left(A^{({v})}-S^{({v})}\right)y)=Tr(y^TL_{x}^{({v})}y).
\end{gather}

We enforce the constraints that view weights are non-negative and the sum of the view weights for all views equals to 1. The view weight $v_i$ for the $i$-th view can be determined as follows:
\begin{gather}
v_i=\frac{1/Tr(y^TL_{x}^{({i})}y)}{\sum_{i=1}^{V}{1/Tr(y^TL_{x}^{({i})}y)}}.
\end{gather}

Therefore, we represent the initial global-view as the fusion matrix $X^f$, which takes into account the relationships between views. The concatenation representation can be expressed as: 
\begin{gather}
X^f=\left[v_1X^{(1)},v_2X^{(2)}\cdots v_VX^{(V)}\right].
\end{gather}
\subsubsection{Global-View Distribution}
Given the observed label matrix $Y$, the complete information could be transformed into distribution that facilitates the reconstruction of the global-view ~\cite{xu2019label}. To achieve this, a nonlinear transformation is applied to the label matrix, mapping the preliminary fusion matrix $X^f$ to a higher-dimensional label space. The higher-dimensional label space $\rho(Y)$ is constructed using Gaussian kernel, which can be expressed as:
\begin{gather}
\rho(Y)=exp(-\frac{J_1Y^T(\sum_{i=1}^{l}{{(y_{ij})}^2)}-YY^T}{{(avg(p\_{dist}))}^2}),
\end{gather}
where $p\_{dist}$ represents the array of pairwise distances in the label matrix, and $avg(p\_{dist})$ means the average value for all pairwise distances. Other definitions such as $J_1\in\mathbb{R}^{n\times 1}$ are represented as the full one matrix, and $Y(\cdot)$ denotes a series of operations on the element in the label matrix. Then, the global-view distribution can be modeled through the coefficient matrix $\hat{{W}_y}$ as:
\begin{gather}
D=\rho(Y){\hat{W}}_y+b=Y_xW_y,
\end{gather}
where $b$ is the bias for this model. To facilitate matrix computations, the coefficient matrix and the higher-dimensional label space are transformed into $W_y=\begin{bmatrix}
\hat{{W}_y}
 \\b
\end{bmatrix}\in \mathbb{R}^{(n+1)\times d}$ and $Y_x=[\rho(Y),J_1]\in \mathbb{R}^{n\times(n+1)}$, respectively. 

Specific to the design for the global-view reconstruction $L_G$, two aspects need to be highlighted. 
First, it is essential to consider the interrelationship between different views. Second, preserving the structure of the global-view is of utmost importance. Since this method is specifically tailored for multi-view scenarios, the roles of the matrices involved in the function $L_G$ cannot be interchanged. Consequently, a more refined regularizer paradigm for $L_G$ can be formulated, taking into account these two aspects: 
\begin{gather}
L_G(\cdot)=\left \| D-X^f \right \|_F^2+Tr(D^TL^YD),
\end{gather}
where the way to define $L^Y$ is the same as $L_{x}^{({v})}$.

\subsection{Uncertainty-Aware Feature Selection}
In terms of dimensionality, feature selection involves selecting meaningful columns. In order to account for the uncertainty of samples, different proportions of information are extracted from each row in the view. By considering both dimensions, the loss function for the uncertainty-aware feature selection model is constructed as:
\begin{gather}
loss= {\textstyle \sum_{i=1}^{V}}\left \| diag(C^{(i)})X^{(i)}W^{(i)}-Y \right \|_F^2+\Phi (W),
\end{gather}
where $W^{(i)}\in\mathbb{R}^{d(i)\times c}$ denotes the feature weight of each view. The $\Phi(W)$ denotes the regularization paradigm of $W$ to control the model complexity, and $W\in\mathbb{R}^{d\times c}$ denotes the global-view feature weight matrix. The inclusion of the learnable confidence vector $C^{(i)}$ effectively mitigates interference, thereby accentuating the significance of pertinent features.

In addition to considering the constraints of view-relationship and structure similarity, the uncertainty perception is also incorporated as a penalty function in this model. Therefore, the penalty function is defined as:
\begin{gather}
penalty= {\textstyle \sum_{i=1}^{V}} \left \| Y_xW_y^{(i)}-diag(C^{(i)})X^{(i)} \right \|_F^2,
\end{gather}
where $W_y=\left \{ {W}_y^{(i)} \right \} _{i=1}^V$ and $W_y^{(i)}\in\mathbb{R}^{(n+1)\times d(i)}$ which is splitting from the coefficient weight $W_y$. Thus, the definition for $Y_xW_y^{(i)}$ can be seen the part of $D$ which can be written as $D^{(i)}$. Simultaneously, each uncertainty-aware view incorporates characteristics of the global distribution, enabling accurate computation of the learnable $C^{(i)}$. Overall, the uncertainty-aware feature selection $L_U$ can be formulated by combining the loss function and the penalty as:
\begin{align}
&L_U\left(\cdot\right)= {\textstyle \sum_{i=1}^{V}}\left \| diag(C^{(i)})X^{(i)}W^{(i)}-Y \right \|_F^2 \notag\\
&+{\textstyle \sum_{i=1}^{V}}\left \|  D^{(i)}-diag(C^{(i)})X^{(i)} \right \|_F^2+\left \| W \right \|_{2,1}.
\end{align}

The term ${diag(C}^{(i)})X^{(i)}$ can be interpreted as the decomposition of the global-view reconstruction into individual views. This is equivalent to taking into account both the sample-level and view-level relationships in the solution for $W^{(i)}$. Therefore, concatenating the feature weights of each view as a variable for controlling model complexity does not affect the feature order. Following the concatenation procedure, we utilize $l_{2,1}$-norm regularization to induce group sparsity ~\cite{yuan2006model}.

\subsection{The Objective Function}
The model for UGRFS is represented by jointly integrating the procedures for global-view distribution and uncertainty-aware feature selection within a unified framework. The detailed objective function is written as follows: 
\begin{align}
&\min_{W^{\left(i\right)},C^{\left(i\right)},W_y^{(i)}} {\textstyle\sum_{i=1}^{V}}\left \| diag(C^{(i)})X^{(i)}W^{(i)}-Y \right \|_F^2 \notag\\
&+\alpha Tr\left(D^TL^YD\right)+\beta{\textstyle \sum_{i=1}^{V}} \left \| D^{(i)}-diag(C^{(i)})X^{(i)} \right \|_F^2 \notag\\
&+\gamma\left \| D-X^f \right \|_F^2+\delta\left \| W \right \|_{2,1},
\end{align}
where $\alpha$, $\beta$, $\gamma$ and $\delta$ are trade-off parameters to keep the balance of the model. It can be seen that the $D$ is associated with three regularizer paradigms, which fuses the information for global structure, sample-confidence and view-relationship. The constraints on $C^{(i)}$ primarily aim to minimize the disparity between the splitting structure of the global-view and the uncertainty-aware view. Due to the comprehensive considerations from $D$ to $C^{(i)}$, accurate mapping between features and labels is achieved, leading to precise feature weights acquisition.

\begin{algorithm}[t]
\caption{Uncertainty-aware Global-view Reconstruction}
\label{alg:algorithm}
\textbf{Input}: Data matrices \bm{$\left \{ {{X^{\left(i\right)}}} \right \} _{i=1}^V$}; Label matrix \bm{$Y$}.\\
\textbf{Parameter}: Parameters $\alpha$, $\beta$, $\gamma$ and $\delta$.\\
\textbf{Output}: Selected features.\\
\begin{algorithmic}[1] 
\STATE Initialize \bm{$W^{\left(i\right)}$}, \bm{$C^{\left(i\right)}$}, \bm{$W_y^{(i)}$};
\REPEAT
\STATE Update the matrix \bm{$W^{\left(i\right)}$} according to Formula (14);
\STATE Update the matrix \bm{$C^{\left(i\right)}$} according to Formula (15);
\STATE Update the matrix \bm{$W_y^{(i)}$} according to Formula (16);
\STATE Update the objective function (13);
\UNTIL{Convergence}
\STATE Obtain the ordered feature sequence by calculating $||\bm{W}_{(\bm{j})}||_{2}$ where $j = 1,2,3,...,d$;
\STATE \textbf{return} Top ranked features as \textit{\textbf{s-UGRFS-f}}.
\end{algorithmic}
\end{algorithm}
\subsection{Optimization}
Considering the smoothness of the objective function, unless otherwise stated, the related definition used in this paper are listed as follows. The Frobenius norm is denoted as $\left \| Y \right \|_{F}=\sqrt{ {\textstyle \sum_{p=1}^{n}} {\textstyle \sum_{q=1}^{l}y_{(pq)}^{2}}}$, and it can be transformed to $\left \| Y \right \|_{F}^2 =Tr(Y^TY)$. The $l_{2,1}$-norm is denoted as $\left \| W \right \|_{2,1}= {\textstyle \sum_{p=1}^{d}} \sqrt{ {\textstyle \sum_{q=1}^{l}}w_{(pq)}^2 }$, and we can relax the corresponding terms as  $\left \| W \right \|_{2,1} =2Tr(W^TEW)$ where the diagonal matrix $e_{ii}=1/(2 \left \| W_i \right \|_{2} )$. The transformation based on Formula (12) can be written as:
\begin{align}
&\Theta =\sum_{i=1}^{V}{Tr({F}^TF)}+\alpha Tr\left(D^TL^YD\right) \notag\\
&+\beta\sum_{i=1}^{V}{Tr({G}^TG)}+\gamma Tr{(D-X^f)}^T(D-X^f) \notag\\
&+2\delta Tr\left(W^TEW\right)-Tr\left(\varphi{W^{\left(i\right)}}^T\right) \notag\\
&-Tr(\psi {W_y^{(i)}}^T)-Tr(\tau {diag(C^{(i)})}^T), 
\end{align}
where $\varphi\in\mathbb{R}^{d(i)\times c}$, $\psi\in\mathbb{R}^{(n+1)\times d(i)}$, $\tau\in\mathbb{R}^{n\times n}$ denote Lagrangian multipliers, and  $F={diag(C}^{\left(i\right)})X^{\left(i\right)}W^{\left(i\right)}-Y$, $G={D^{(i)}-diag(C}^{(i)})X^{(i)}$. To solve the minimization for Formula (13), we take advantage of a non-negative alternating iterative optimization algorithm where $A^{(i)}={diag(C}^{\left(i\right)})X^{\left(i\right)}$. The description of the procedure is presented in Algorithm 1.
\begin{gather}
W^{\left(i\right)}\gets W^{\left(i\right)}\circ\frac{{A^{(i)}}^TY}{{A^{(i)}}^TA^{(i)}W^{\left(i\right)}+2\delta EW^{\left(i\right)}}.
\end{gather}

\begin{gather}
C^{\left(i\right)}\gets C^{\left(i\right)}\circ\frac{Y{W^{\left(i\right)}}^T{X^{\left(i\right)}}^T+\beta D^{(i)}{X^{(i)}}^T}{A^{(i)}W^{\left(i\right)}{W^{\left(i\right)}}^T{X^{\left(i\right)}}^T+\beta A^{(i)}{X^{\left(i\right)}}^T}.
\end{gather}

\begin{gather}
W_y^{(i)}\gets W_y^{(i)}\circ\frac{\alpha{Y_x}^TS^YD^{(i)}+\beta{Y_x}^T{A^{(i)}}+\gamma{Y_x}^TX^{f\left(i\right)}}{\alpha{Y_x}^TA^YD^{(i)}+(\beta+\gamma){Y_x}^TD^{(i)}}.
\end{gather}

\begin{figure}[tb]
 \begin{minipage}{0.49\linewidth}
 	\vspace{3pt}
 	\centerline{\includegraphics[width=\textwidth]{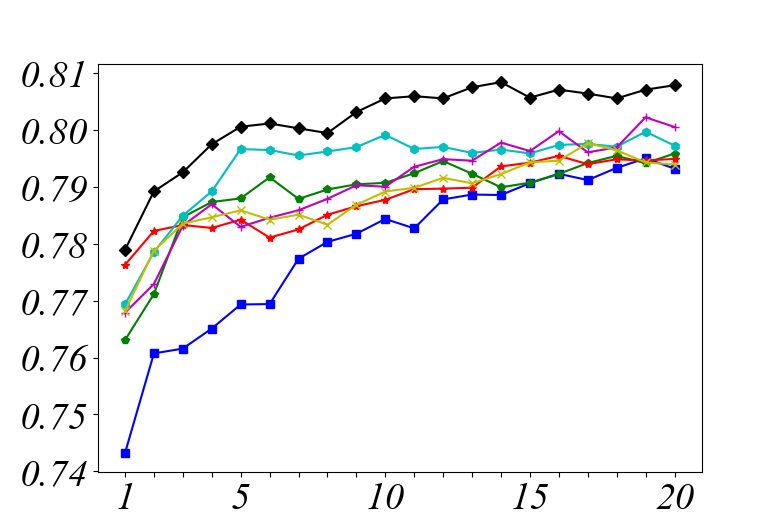}}
\centerline{(a) {Average Precision}}
 	\vspace{3pt}
	\centerline{\includegraphics[width=\textwidth]{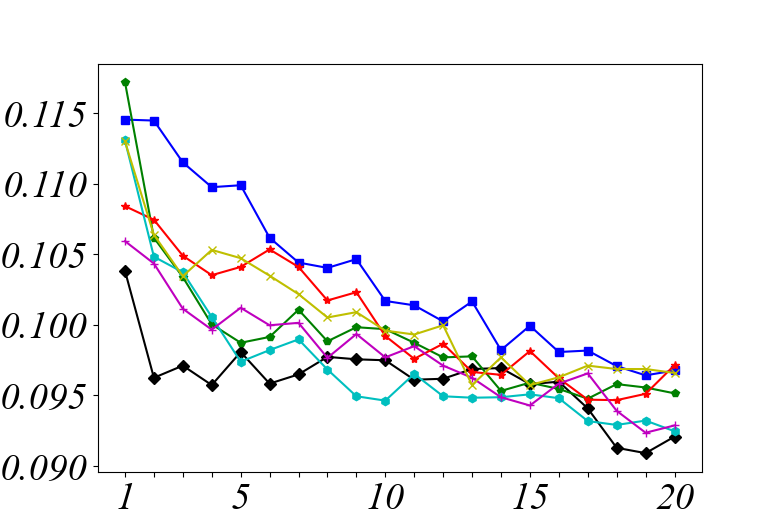}}
\centerline{(c) {Hamming Loss}}
 	\vspace{3pt}
 	
 \end{minipage}
 \begin{minipage}{0.49\linewidth}
	\vspace{3pt}
	\centerline{\includegraphics[width=\textwidth]{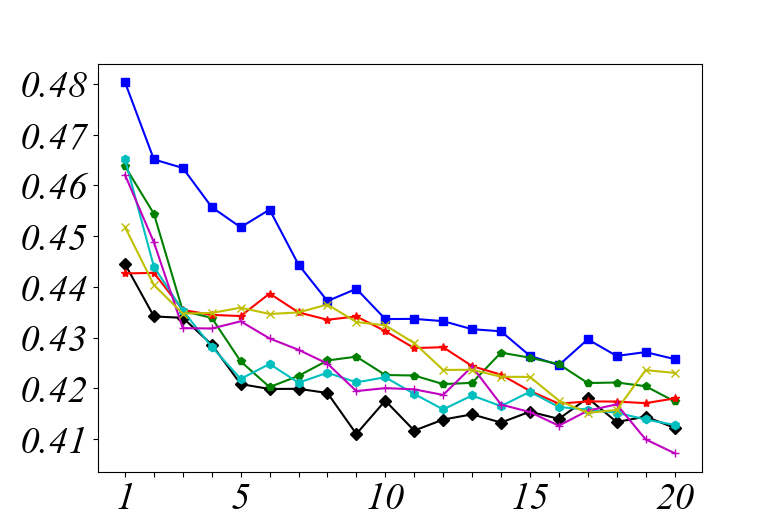}}
\centerline{(b) {Coverage}}
	\vspace{3pt}
	\centerline{\includegraphics[width=\textwidth]{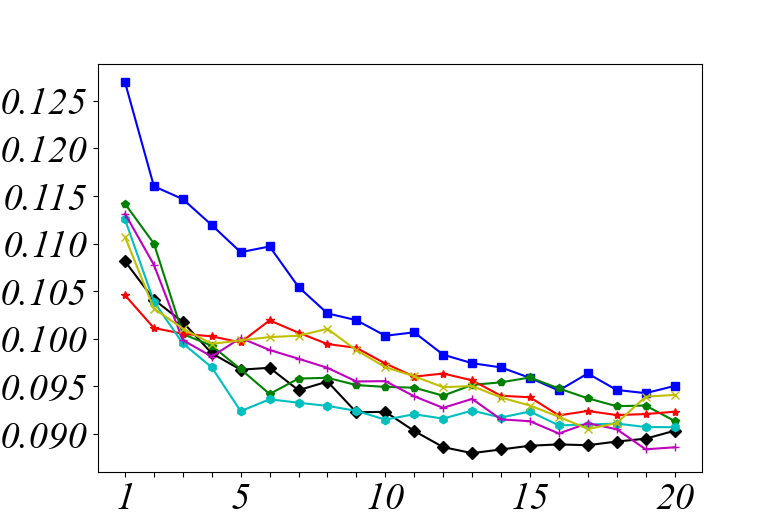}}
\centerline{(d) {Ranking Loss}}
 	\vspace{3pt}
  \end{minipage}
 
  \caption{Seven methods on SCENE in terms of Average Precision, Coverage, Hamming Loss and Ranking loss.}
  \label{fig2}
\vspace{-10pt}
\end{figure}

\subsection{Complexity Analysis}
The optimization procedure consists of three primary components, assuming a feature matrix dimension of $d$ for each view. When updating $W^{\left(i\right)}$, the computation complexity is $O(d^2n)$. The computation complexity for updating $C^{\left(i\right)}$ is also $O(n^2d)$. Updating $W_y^{(i)}$ has a computation complexity of $O(n^3+n^2d)$. Considering these complexities, the entire training procedure can be conservatively approximated as $O(n^3+n^2d+d^2n)$ per iteration.

\section{Experiments}
\subsection{Experimental Setup}
\subsubsection{Datasets}
\begin{table*}[htbp]

	\begin{center}

		\footnotesize
		\begin{tabular}{|p{1.45cm}|p{0.8cm}|p{1.4cm}|p{1.2cm}|p{1.2cm}|p{1.7cm}|p{1.85cm}|}
			\hline
			Views  & yeast  & SCENE& VOC07& MIRFlickr & IAPRTC12&3Sources\\
			\hline
			View-1($d_1$)&GE(79)&CH(64)&DH(100)&DH(100)&DH(100)&BBC(1000)\\
			View-2($d_2$)	&PP(24)	&CM(225)&GIST(512)&GIST(512)&DHV3H1(300)&Reuters(1000)\\
			View-3($d_3$)&-	&CORR(144)	&HH(100) 	&HH(100)&GIST(512)&Guardian(1000)\\
			View-4($d_4$)&-			&	EDH(73)	&-&-&HHV3H1(300)&-\\
			View-5($d_5$)&-&		WT(128)&-&-	&HH(100)&-\\
			\hline
			Instances($n$)&2417&	4400&3817&	4053&	4999&169\\
			Features($d$)&103&	634&712&712&1312&3000\\
			Labels($l$)&14&	33&20&	38&	260&6\\
			\hline
		\end{tabular}
\caption{The detailed information for the datasets in our experiments.}\label{table 3} 
	\end{center}
\vspace{-10pt}
\end{table*}

As shown in Table 1, we evaluate our method on six widely used multi-view multi-label datasets, namely yeast ~\cite{elisseeff2001kernel}, SCENE ~\cite{chua2009nus}, VOC07 ~\cite{everingham2010pascal}, MIRFlickr ~\cite{huiskes2008mir}, IAPRTC12 ~\cite{escalante2010segmented}, and 3Sources ~\cite{greene2009matrix}. These datasets cover various domains including genes, vision, and news, and have different numbers of views ranging from 2 to 5. The number of features varies from a few hundred to several thousand. In cases where a view is absent, it is indicated by `-'.

\subsubsection{Comparing Methods}
Comparison methods include two multi-view multi-label methods, i.e., M2LD(`blue') ~\cite{liu2023label}, and MSFS(`green') ~\cite{zhang2020multi}, and four multi-label methods, i.e., MoRE(`red') ~\cite{liu2022more}, MRDM(`cyan') ~\cite{huang2021multi}, MIFS(`pink') ~\cite{jian2016multi} and CLML(`yellow') ~\cite{li2022learning}. These methods share a common emphasis on embedded learning, which primarily focuses on the feature space. Thus, the feature weights could be acquired for sorting.

\subsubsection{Evaluation Metrics}
We evaluate the effectiveness of our method using four widely adopted metrics ~\cite{zhang2013review,gibaja2015tutorial}, i.e., Average Precision (AP), Coverage, Hamming Loss (HL) and Ranking Loss (RL). A higher value of AP indicates better performance, while the remaining three metrics operate in the opposite direction. The evaluation results are presented as mean accuracy along with standard deviation obtained from five-fold cross-validation. The parameters for each method are tuned within the range of $\left \{ {10}^{-3},\ {10}^{-2}, ..., {10}^3 \right \}$. These methods adhere to the above standards to ensure the validity of the comparative results presented in this paper.

\subsection{Experimental Results}
\subsubsection{Feature Selection Performance}

\begin{figure}[tb]
 \begin{minipage}{0.48\linewidth}
 	\vspace{3pt}
 	\centerline{\includegraphics[width=\textwidth]{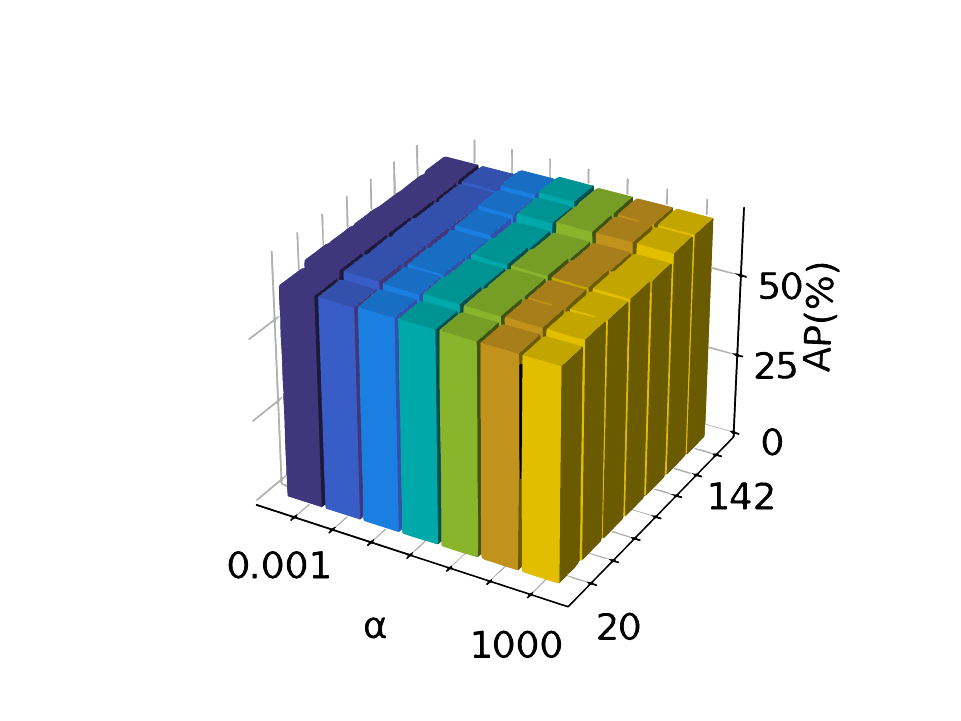}}

 	\vspace{3pt}
	\centerline{\includegraphics[width=\textwidth]{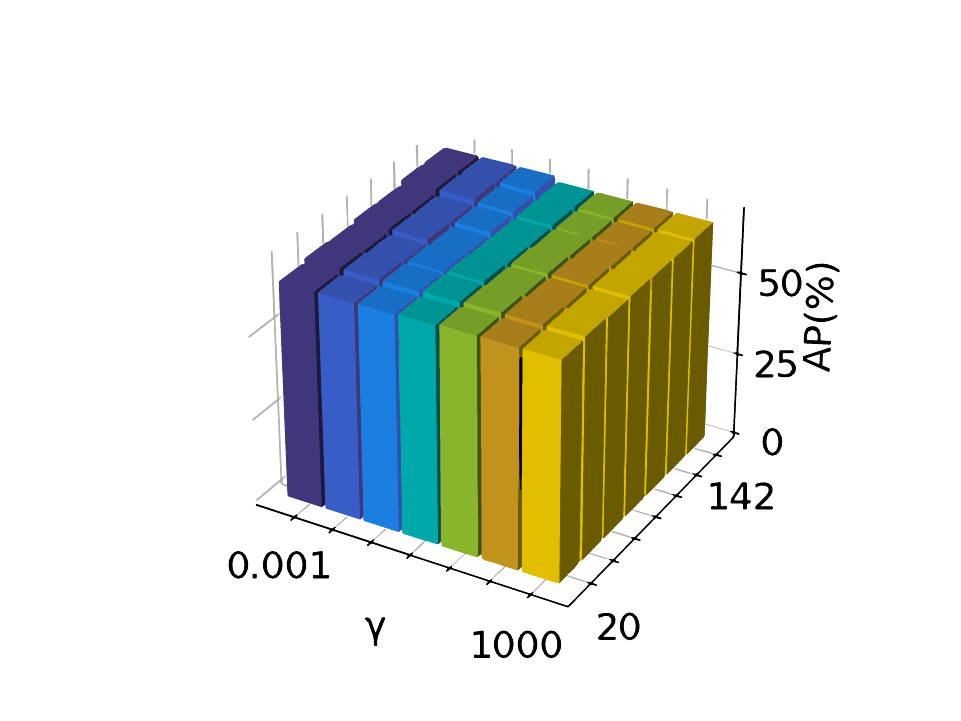}}

 	\vspace{3pt}
 	
 \end{minipage}
 \begin{minipage}{0.48\linewidth}
	\vspace{3pt}
	\centerline{\includegraphics[width=\textwidth]{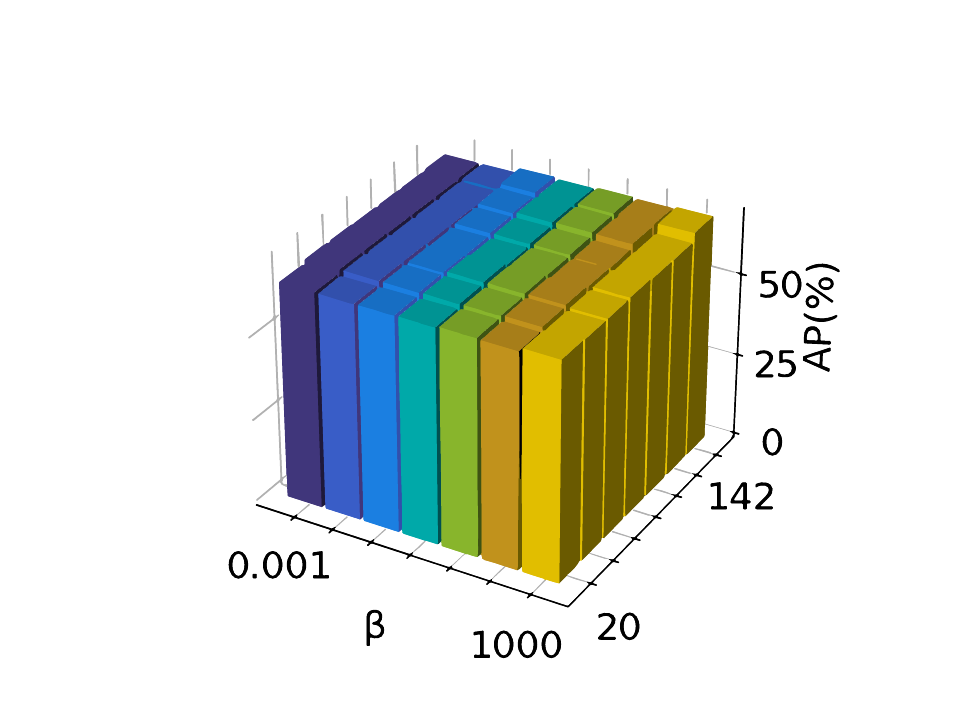}}

	\vspace{3pt}
	\centerline{\includegraphics[width=\textwidth]{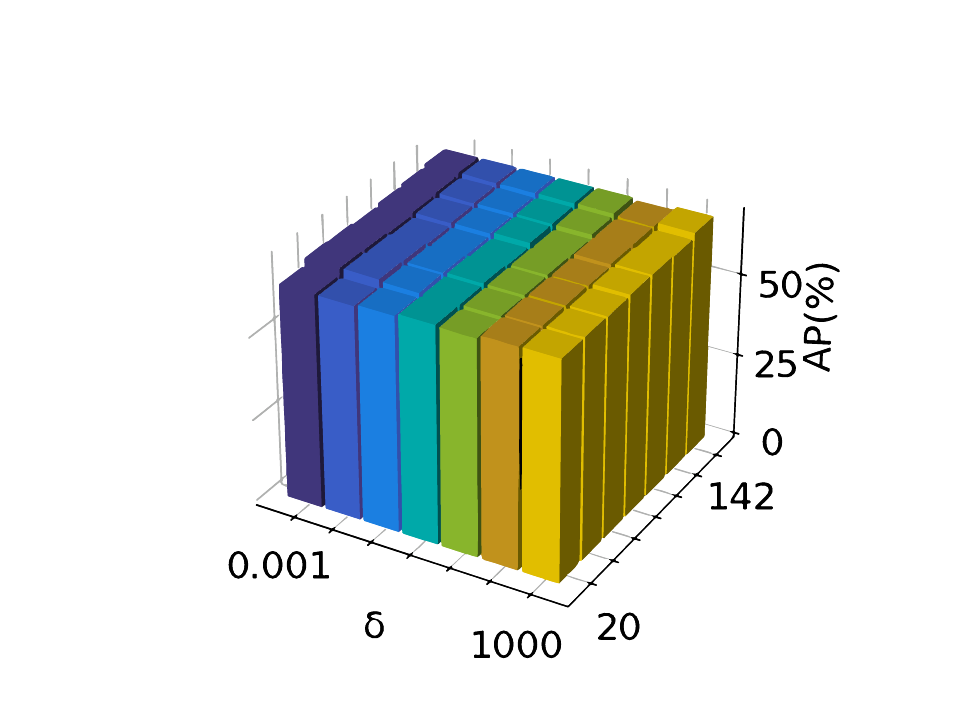}}

 	\vspace{3pt}

  \end{minipage}
 
  \caption{Parameter sensitivity studies on the MIRFlickr dataset.}
  \label{fig3}
\vspace{-10pt}
\end{figure}

\begin{figure}[h]
 \begin{minipage}{0.4\linewidth}
 	\vspace{3pt}
 	\centerline{\includegraphics[width=\textwidth]{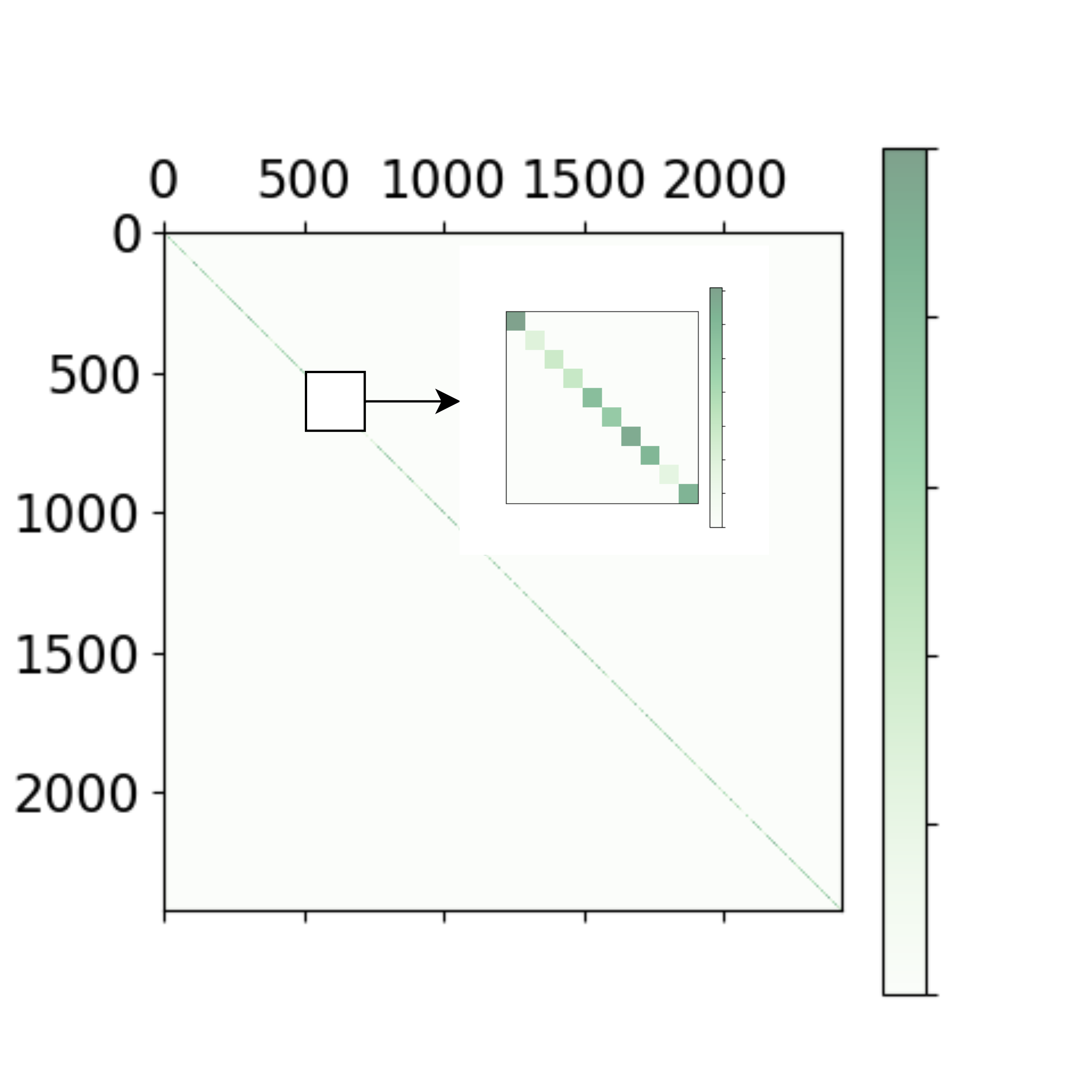}}
\centerline{(a)}
 	\vspace{3pt}

 \end{minipage}
 \begin{minipage}{0.6\linewidth}
	\vspace{3pt}
	\centerline{\includegraphics[width=\textwidth]{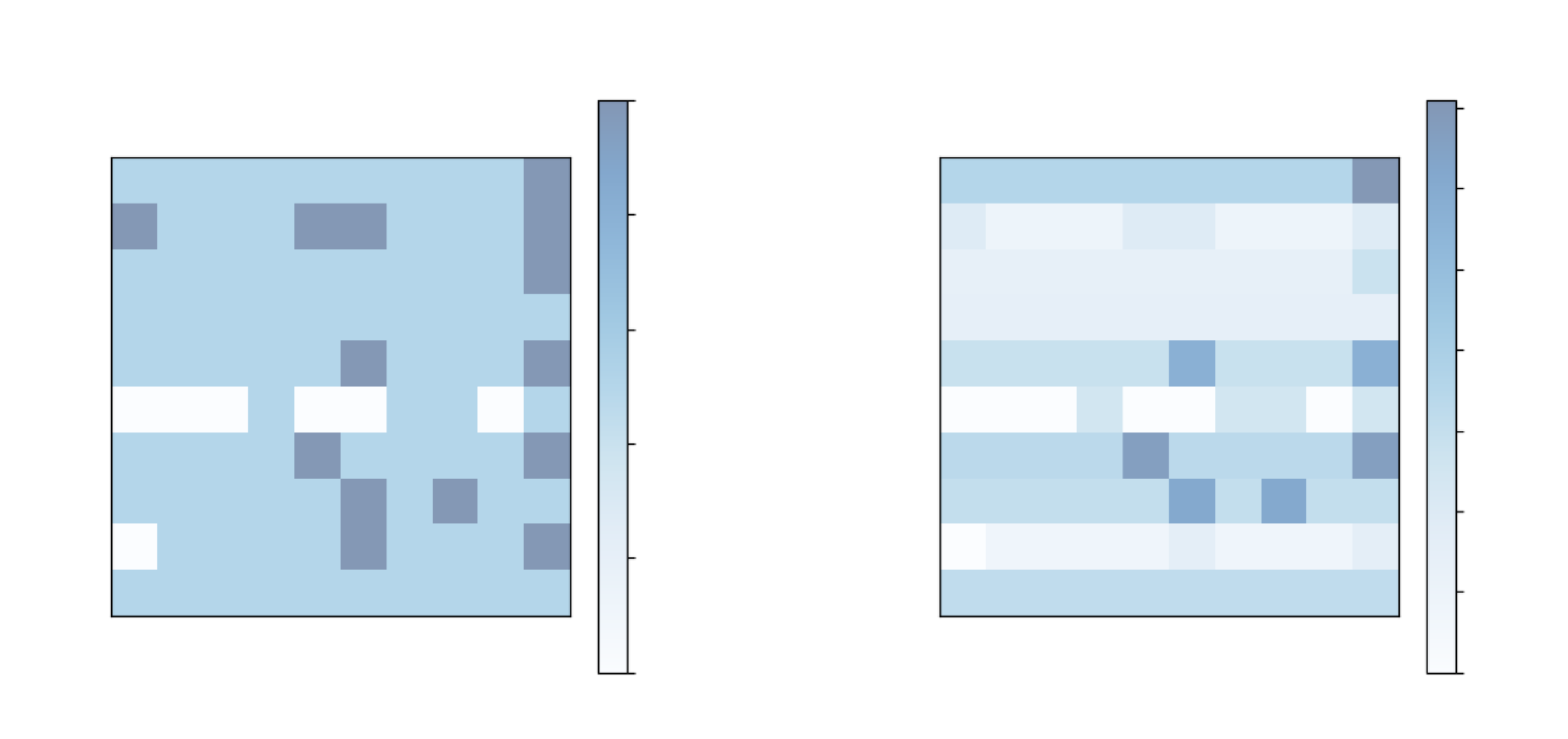}}
\centerline{(b)}
	\vspace{3pt}
	
  \end{minipage}

  \caption{Heatmap of the (a) sample confidence and (b) original features and uncertainty-aware features on yeast.}
  \label{fig4}
\vspace{-10pt}
\end{figure}

\begin{table*}[tb]
\small
    \renewcommand{\arraystretch}{1.2}
    
    \label{tab:my-table}
    \centering
    \setlength{\tabcolsep}{0.4mm}{
   \begin{tabular*}{\hsize}{@{\extracolsep{\fill}}c| c c c c c c c}
        \toprule
        Dataset &UGRFS&M2LD&MSFS&MoRE&MRDM&MIFS&CLML\\
	   \midrule
        \multicolumn{8}{c}{ \textit{AP} $\uparrow$}\\
        \midrule
        yeast&	\textbf{0.6725±0.014}&	0.6587±0.015	&0.6495±0.012&	0.6552±0.016&	\underline{0.6595±0.016}&0.6445±0.014&	0.6496±0.012\\
 \hline
      SCENE&	\textbf{0.8010±0.009}&	0.7792±0.014&	0.7887±0.008	&0.7877±0.007&	\underline{0.7929±0.010}&0.7896±0.010&	0.7870±0.008\\
        \hline
        VOC07&	\textbf{0.5871±0.010}&	0.5763±0.009&	0.5751±0.006&	0.5664±0.016&	\underline{0.5836±0.008}&	0.5690±0.010&	0.5790±0.011\\
\hline
        MIRFlickr&	\textbf{0.6770±0.007}&	0.6436±0.013&	0.6417±0.010&	0.6322±0.011&	0.6407±0.008&0.6398±0.009&\underline{	0.6442±0.007}\\
       
        \hline
IAPRTC12&	\textbf{0.1474±0.003}&	0.1370±0.006&	0.1385±0.005&	0.1263±0.006&	0.1400±0.005&\underline{0.1416±0.003}&	0.1352±0.002\\
\hline
3Sources&	\textbf{0.4728±0.055}&	0.4368±0.038&	0.4252±0.035&	-&	0.4196±0.041&\underline{0.4623±0.035}	&0.4232±0.046\\

 \midrule
        \multicolumn{8}{c}{ \textit{Coverage} $\downarrow$}\\
        \midrule

       yeast&	\textbf{0.6239±0.022}&	0.6425±0.017&	0.6375±0.021&	0.6357±0.022&	0.6426±0.023&\underline{0.6339±0.029}	&0.6407±0.019\\
       \hline
      SCENE&	\textbf{0.4214±0.014}&	0.4416±0.016&	0.4271±0.012&		0.4295±0.010&	\underline{0.4252±0.017}&	\underline{0.4252±0.014}&	0.4308±0.012\\
        \hline
VOC07&	\textbf{0.4425±0.016}&	0.4547±0.015&	\underline{0.4484±0.016}&	0.4763±0.027&	0.4531±0.014&	0.4503±0.022&	0.4569±0.015\\
\hline
       MIRFlickr&	\textbf{0.5885±0.007}&	\underline{0.6069±0.009}&	0.6173±0.013&	0.6397±0.010&	0.6128±0.017&0.6325±0.013	&0.6191±0.015\\
        
        \hline
IAPRTC12&	\textbf{0.4996±0.008}&	0.5131±0.011&	0.5098±0.009&	0.5877±0.020&	0.5097±0.017&\underline{0.5004±0.007}	&0.5149±0.005\\
 \hline
        3Sources	&\textbf{0.5301±0.039}&0.6397±0.030&	0.6602±0.037& -&	0.6723±0.042&\underline{0.6310±0.031}	&0.6727±0.046 \\

        \bottomrule
    \end{tabular*}}
\caption{Experimental results (mean ± std) in terms of AP and Coverage, where the 1st/2nd best results are shown in boldface/underline.}
\vspace{-10pt}
\end{table*}

\begin{table*}[tb]
\small
    \renewcommand{\arraystretch}{1.2}
    
    \label{tab:my-table}
    \centering
    \setlength{\tabcolsep}{0.4mm}{
   \begin{tabular*}{\hsize}{@{\extracolsep{\fill}}c| c c c c c c c}
        \toprule
       Dataset &UGRFS&M2LD&MSFS&MoRE&MRDM&MIFS&CLML\\
	   \midrule
        \multicolumn{8}{c}{ \textit{HL} $\downarrow$}\\
        \midrule

         yeast&\textbf{	0.2257±0.005}	&0.2341±0.002&	0.2299±0.004&\underline{	0.2259±0.006}&	0.2358±0.007&0.2279±0.003	&0.2310±0.004\\
         \hline
      SCENE&	\textbf{0.0978±0.004}&	0.1040±0.006&	0.0999±0.005	&	0.1007±0.005	&\underline{0.0979±0.005}&0.0987±0.004	&0.1010±0.005\\
        \hline
VOC07&	\textbf{0.0847±0.001}&	\underline{0.0862±0.002}&	0.0893±0.001&	0.0890±0.001&	0.0880±0.001&	0.0886±0.001&	0.0884±0.001\\
\hline
        MIRFlickr&	\textbf{0.1775±0.005}	&0.2029±0.007&	0.1944±0.002	&0.1931±0.003&	0.1946±0.003&0.1925±0.004&\underline{0.1917±0.003}\\
       
        \hline
IAPRTC12&	\textbf{0.01496±0.00}&	0.01653±0.00	&0.01561±0.00&\underline{0.01550±0.00	}&0.01560±0.00&0.01560±0.00	&0.01554±0.00\\
\hline
        3Sources&	\textbf{0.2097±0.020}&	0.2491±0.025&	0.2357±0.017&	-	&0.2358±0.017	&\underline{0.2332±0.020}&0.2471±0.018\\

 \midrule
        \multicolumn{8}{c}{ \textit{RL} $\downarrow$}\\
        \midrule

        yeast	&\textbf{0.2480±0.010}&	0.2605±0.012&	0.2614±0.010&		\underline{0.2563±0.017}&	0.2574±0.015&0.2661±0.015	&0.2628±0.010\\
       \hline
      SCENE&	\textbf{0.0942±0.007}&	0.1037±0.009&	0.0970±0.006&	0.0974±0.005&	\underline{0.0949±0.007}&0.0962±0.007	&0.0980±0.006\\
        \hline
VOC07&	\textbf{0.2111±0.006}&	0.2142±0.010&	0.2133±0.010&	0.2298±0.017&	\underline{0.2124±0.009}&	0.2162±0.014&	0.2157±0.011\\
\hline
      MIRFlickr	&\textbf{0.1629±0.005}&\underline{0.1672±0.008}&	0.1855±0.006&	0.1925±0.006&	0.1858±0.007&0.1880±0.006	&0.1844±0.006\\
        
        \hline
IAPRTC12&	\textbf{0.2020±0.005}&	0.2070±0.009	&0.2093±0.006&	0.2452±0.010&	0.2069±0.013&\underline{0.2028±0.005}	&0.2134±0.003\\
 \hline
       3Sources	&\textbf{0.4137±0.049}	&0.5045±0.039&	0.5253±0.043&	-&	0.5360±0.051&\underline{0.4830±0.037}	&0.5345±0.055\\
      
        \bottomrule
    \end{tabular*}}
\caption{Experimental results (mean ± std) in terms of HL and RL, where the 1st/2nd best results are shown in boldface/underline.}

\end{table*}

\begin{table}[htbp]

	\begin{center}
\footnotesize
		
		\begin{tabular}{p{1.2cm}p{0.9cm}p{0.8cm}p{1cm}p{1cm}p{1cm}}
			\toprule[1.5pt]
			&Metric&UGRFS & UGRFS$_{v1}$ & UGRFS$_{v2}$ & UGRFS$_{v3}$ \\
			\hline
			yeast&AP&0.6725&0.6616&0.6603&0.6610\\
			&Coverage&0.6239&0.6354		&0.6245	&0.6237	\\
			&RL&0.2480&	0.2565&	0.2496&	0.2516\\
\hline
			MIRFLickr&AP&0.6770&0.6646	&0.6729&0.6470	\\
			&Coverage&0.5885&0.6058&	0.5924&	0.6249\\
			&RL&0.1629&	0.1713&0.1659	&	0.1849\\
\hline
			IAPRTC12&AP&0.1474&0.1471&		0.1472&0.1406	\\
			&Coverage&0.4996&0.5332&	0.5358&	0.5022\\
			&RL&0.2020&	0.2034&0.2076	&	0.2073\\

			\bottomrule[1.5pt]
		\end{tabular}
\caption{Ablation studies on the yeast, MIRFlickr and IAPRTC12 datasets for Average Precision, Coverage and Ranking loss.}\label{table 6} 
	\end{center}
\vspace{-20pt}
\end{table}

Table 2-3 shows the performance of UGRFS, compared with other methods. Each value is calculated based on results obtained using one to twenty percentage features. As can be seen, our method demonstrates superior performance compared to all baseline methods on six datasets. Although the performance of our method is not significant different from some second-ranking methods, there remains a substantial gap of 4.84\% on AP compared to the second-ranking method. The gap can increase to 14.3\% when comparing to the last-ranking method. Upon closer analysis, we observe that MIFS and MRDM achieve better performance, ranking second in 37.5\% and 29.17\% of cases, respectively. The effective combination of feature and label information proves crucial, especially in the multi-view scenario when comparing MIFS and MSFS. We also provide a clear illustration of our performance for all metrics on one dataset in Figure 2.

\subsubsection{Parameter Analysis}

As four parameters $\alpha$, $\beta$, $\gamma$ and $\delta$ of UGRFS are involved in two parts, including global-view reconstruction and sparsity in row. We individually adjust each parameter while keeping the others fixed. Figure 3 illustrates the tuning range on the X-axis, ranging from 0.001 to 1000, while the number of features is represented on the Y-axis. The plot demonstrates the impact of these four parameters on our method. Overall, our method exhibits relatively low sensitivity to these parameters, with weak fluctuations primarily observed as the number of features increases.

\subsubsection{Ablation Study}

To verify the effectiveness of our design for the objective function, we propose three types of transformations for the original function. The first transformation, UGRFS$_{v1}$, focuses on evaluating the impact of sample confidence, which is beneficial for uncertainty-aware feature selection. The remaining two transformations, UGRFS$_{v2}$ and UGRFS$_{v3}$, prioritize the impact of global-view reconstruction. The main difference between them is that UGRFS$_{v2}$ disregards the variation between views in the initial global-view, while UGRFS$_{v3}$ excludes the related regularizer paradigm for global-view reconstruction, setting parameters $\alpha$, $\beta$ and $\gamma$ as 0. Table 4 presents the performance of the three transformations compared to the original method in terms of AP, Coverage, and RL on yeast, MIRFlickr, and IAPRTC12. UGRFS achieves the best overall performance, except for a minor difference in terms of Coverage on the yeast dataset. UGRFS$_{v3}$ exhibits the lowest performance in terms of AP, as it ignores all reconstructed components. However, UGRFS$_{v2}$ occasionally performs worse than UGRFS$_{v3}$. This finding emphasizes the importance of correctly computation of view weights. In conclusion, each component within the objective function of our method holds irreplaceable significance.

\subsubsection{Visualization of Sample Confidence}
To further visualize the variation in sample confidence, we utilize heatmaps to illustrate the differences between samples, as shown in Figure 4a. It is evident that some samples exhibit similar confidence. But individual sample confidence is not exactly equal due to factors such as noise. In Figure 4b, the distinction lies in the inclusion or exclusion of sample confidence. This results in certain features becoming uninformative while a few features remain significant. Consequently, these changes alter the graph structure of the features.

\subsubsection{Convergence}
\begin{figure}[tb]
 \begin{minipage}{0.49\linewidth}
 	\vspace{-10pt}
 	\centerline{\includegraphics[width=\textwidth]{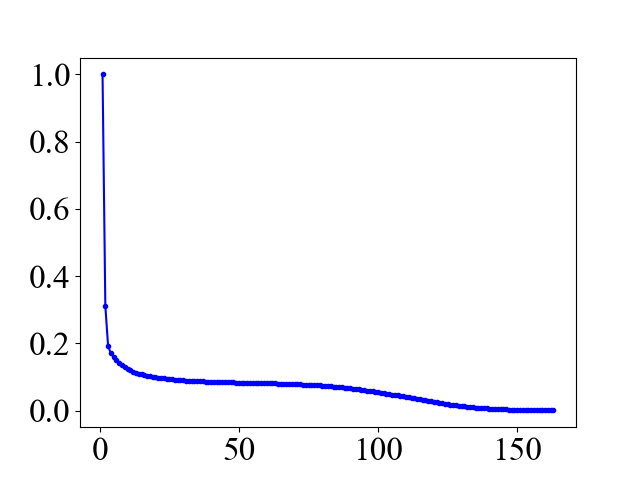}}
\centerline{(a) {MIRFlickr}}
 	\vspace{-3pt}

 \end{minipage}
 \begin{minipage}{0.49\linewidth}
	\vspace{-10pt}
	\centerline{\includegraphics[width=\textwidth]{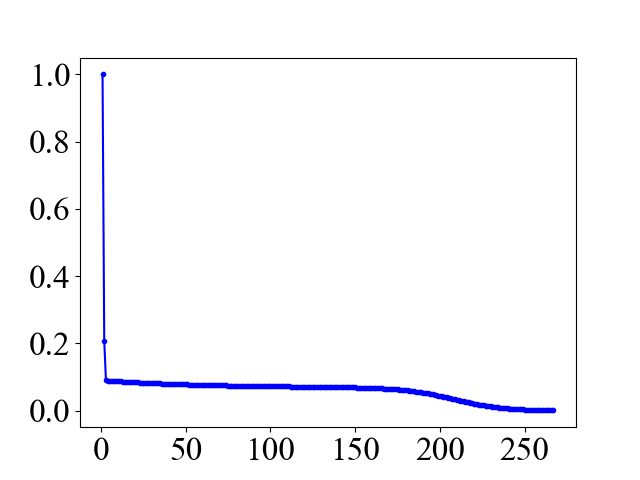}}
\centerline{(b) {VOC07}}
	\vspace{-3pt}
	
  \end{minipage}

  \caption{Convergence curves analysis of UGRFS on (a) MIRFlickr and (b) VOC07.}
  \label{fig5}
\vspace{10pt}
\end{figure}
In Figure 5, we represent the Y-axis as the value of $(z^{t-1}-z^{t})/z^{t-1}$, where $z^t$ denotes the objective values at time t. The X-axis denotes the number of iterations, and the stopping criterion depends on whether the value is small, such as 0.001. Additionally, the setting for Y-axis could also reflect the speed of the convergence. To facilitate observation of the oscillation, we set the initial point to approximately one. It can be seen that the curve rapidly declines in the initial iterations without any protrusions. 

\section{Conclusion}
This paper explores the consistency and complementary in MVML from a novel perspective. By leveraging the label matrix, it reconstructs the global-view, taking into account the graph structure similarity, sample confidence, and view-relationship to preserve informative features and samples. Extensive experiments demonstrate its stability and the superior performance in feature selection. Moving forward, we aim to further investigate valuable methods for view reconstruction.

\section{Acknowledgments}
This work is funded by: by Science Foundation of Jilin Province of China under Grant No. 20230508179RC, and China Postdoctoral Science Foundation funded project under Grant No. 2023M731281,  and Changchun Science and Technology Bureau Project 23YQ05.

\bibliography{aaai25}

\begin{thebibliography}{54}
\providecommand{\natexlab}[1]{#1}

\bibitem[{Bharati, Mondal, and Podder(2023)}]{bharati2023review}
Bharati, S.; Mondal, M. R.~H.; and Podder, P. 2023.
\newblock A Review on Explainable Artificial Intelligence for Healthcare: Why,
  How, and When?
\newblock \emph{IEEE Transactions on Artificial Intelligence}.

\bibitem[{Chua et~al.(2009)Chua, Tang, Hong, Li, Luo, and Zheng}]{chua2009nus}
Chua, T.-S.; Tang, J.; Hong, R.; Li, H.; Luo, Z.; and Zheng, Y. 2009.
\newblock Nus-wide: a real-world web image database from national university of
  singapore.
\newblock In \emph{Proceedings of the ACM international conference on image and
  video retrieval}, 1--9.

\bibitem[{Chung(1997)}]{chung1997spectral}
Chung, F.~R. 1997.
\newblock \emph{Spectral graph theory}, volume~92.
\newblock American Mathematical Soc.

\bibitem[{Cohen et~al.(2023)Cohen, Shnitzer, Kluger, and Talmon}]{cohen2023few}
Cohen, D.; Shnitzer, T.; Kluger, Y.; and Talmon, R. 2023.
\newblock Few-sample feature selection via feature manifold learning.
\newblock In \emph{International Conference on Machine Learning}, 6296--6319.
  PMLR.

\bibitem[{Elisseeff and Weston(2001)}]{elisseeff2001kernel}
Elisseeff, A.; and Weston, J. 2001.
\newblock A kernel method for multi-labelled classification.
\newblock \emph{Advances in neural information processing systems}, 14.

\bibitem[{Escalante et~al.(2010)Escalante, Hern{\'a}ndez, Gonzalez,
  L{\'o}pez-L{\'o}pez, Montes, Morales, Sucar, Villasenor, and
  Grubinger}]{escalante2010segmented}
Escalante, H.~J.; Hern{\'a}ndez, C.~A.; Gonzalez, J.~A.; L{\'o}pez-L{\'o}pez,
  A.; Montes, M.; Morales, E.~F.; Sucar, L.~E.; Villasenor, L.; and Grubinger,
  M. 2010.
\newblock The segmented and annotated IAPR TC-12 benchmark.
\newblock \emph{Computer vision and image understanding}, 114(4): 419--428.

\bibitem[{Everingham and Winn(2010)}]{everingham2010pascal}
Everingham, M.; and Winn, J. 2010.
\newblock The PASCAL visual object classes challenge 2007 (VOC2007) development
  kit.
\newblock \emph{Int. J. Comput. Vis}, 88(2): 303--338.

\bibitem[{Fan et~al.(2021{\natexlab{a}})Fan, Liu, Liu, Du, Lan, and
  Wu}]{fan2021manifold}
Fan, Y.; Liu, J.; Liu, P.; Du, Y.; Lan, W.; and Wu, S. 2021{\natexlab{a}}.
\newblock Manifold learning with structured subspace for multi-label feature
  selection.
\newblock \emph{Pattern Recognition}, 120: 108169.

\bibitem[{Fan et~al.(2021{\natexlab{b}})Fan, Liu, Weng, Chen, Chen, and
  Wu}]{fan2021multi}
Fan, Y.; Liu, J.; Weng, W.; Chen, B.; Chen, Y.; and Wu, S. 2021{\natexlab{b}}.
\newblock Multi-label feature selection with local discriminant model and label
  correlations.
\newblock \emph{Neurocomputing}, 442: 98--115.

\bibitem[{Gibaja and Ventura(2015)}]{gibaja2015tutorial}
Gibaja, E.; and Ventura, S. 2015.
\newblock A tutorial on multilabel learning.
\newblock \emph{ACM Computing Surveys (CSUR)}, 47(3): 1--38.

\bibitem[{Greene and Cunningham(2009)}]{greene2009matrix}
Greene, D.; and Cunningham, P. 2009.
\newblock A matrix factorization approach for integrating multiple data views.
\newblock In \emph{Joint European conference on machine learning and knowledge
  discovery in databases}, 423--438. Springer.

\bibitem[{Han et~al.(2022)Han, Yang, Huang, Zhang, and Yao}]{han2022multimodal}
Han, Z.; Yang, F.; Huang, J.; Zhang, C.; and Yao, J. 2022.
\newblock Multimodal dynamics: Dynamical fusion for trustworthy multimodal
  classification.
\newblock In \emph{Proceedings of the IEEE/CVF Conference on Computer Vision
  and Pattern Recognition}, 20707--20717.

\bibitem[{Huang et~al.(2019)Huang, Qin, Zheng, Cheng, Yuan, Zhang, and
  Huang}]{huang2019improving}
Huang, J.; Qin, F.; Zheng, X.; Cheng, Z.; Yuan, Z.; Zhang, W.; and Huang, Q.
  2019.
\newblock Improving multi-label classification with missing labels by learning
  label-specific features.
\newblock \emph{Information Sciences}, 492: 124--146.

\bibitem[{Huang and Wu(2021)}]{huang2021multi}
Huang, R.; and Wu, Z. 2021.
\newblock Multi-label feature selection via manifold regularization and
  dependence maximization.
\newblock \emph{Pattern Recognition}, 120: 108149.

\bibitem[{Huiskes and Lew(2008)}]{huiskes2008mir}
Huiskes, M.~J.; and Lew, M.~S. 2008.
\newblock The mir flickr retrieval evaluation.
\newblock In \emph{Proceedings of the 1st ACM international conference on
  Multimedia information retrieval}, 39--43.

\bibitem[{Jian et~al.(2016)Jian, Li, Shu, and Liu}]{jian2016multi}
Jian, L.; Li, J.; Shu, K.; and Liu, H. 2016.
\newblock Multi-label informed feature selection.
\newblock In \emph{IJCAI}, volume~16, 1627--33.

\bibitem[{Jiang et~al.(2024)Jiang, Wu, Zhou, Cohn, Liu, Sheng, and
  Chen}]{jiang2024multiview}
Jiang, B.; Wu, X.; Zhou, X.; Cohn, A.~G.; Liu, Y.; Sheng, W.; and Chen, H.
  2024.
\newblock Semi-Supervised Multi-View Feature Selection with Adaptive Graph
  Learning.
\newblock \emph{IEEE Transactions on Neural Networks and Learning Systems},
  35(3): 3615--3629.

\bibitem[{Klonecki, Teisseyre, and Lee(2023)}]{klonecki2023cost}
Klonecki, T.; Teisseyre, P.; and Lee, J. 2023.
\newblock Cost-constrained feature selection in multilabel classification using
  an information-theoretic approach.
\newblock \emph{Pattern Recognition}, 141: 109605.

\bibitem[{Li et~al.(2022)Li, Li, Hu, and Yu}]{li2022learning}
Li, J.; Li, P.; Hu, X.; and Yu, K. 2022.
\newblock Learning common and label-specific features for multi-Label
  classification with correlation information.
\newblock \emph{Pattern Recognition}, 121: 108259.

\bibitem[{Li and Chen(2021)}]{li2021concise}
Li, X.; and Chen, S. 2021.
\newblock A concise yet effective model for non-aligned incomplete multi-view
  and missing multi-label learning.
\newblock \emph{IEEE Transactions on Pattern Analysis and Machine
  Intelligence}, 44(10): 5918--5932.

\bibitem[{Li, Hu, and Gao(2023)}]{li2023multi}
Li, Y.; Hu, L.; and Gao, W. 2023.
\newblock Multi-label feature selection via robust flexible sparse
  regularization.
\newblock \emph{Pattern Recognition}, 134: 109074.

\bibitem[{Lin et~al.(2023)Lin, He, Guo, and Ding}]{lin2023multi}
Lin, Y.; He, Z.; Guo, L.; and Ding, W. 2023.
\newblock Multi-Label Feature Selection via Positive or Negative Correlation.
\newblock \emph{IEEE Transactions on Emerging Topics in Computational
  Intelligence}.

\bibitem[{Liu et~al.(2023{\natexlab{a}})Liu, Li, Xiao, Chen, Liu, Liu, Wang,
  and Sun}]{liu2023multi}
Liu, B.; Li, W.; Xiao, Y.; Chen, X.; Liu, L.; Liu, C.; Wang, K.; and Sun, P.
  2023{\natexlab{a}}.
\newblock Multi-view multi-label learning with high-order label correlation.
\newblock \emph{Information Sciences}, 624: 165--184.

\bibitem[{Liu et~al.(2023{\natexlab{b}})Liu, Wen, Luo, Huang, Wu, and
  Xu}]{liu2023dicnet}
Liu, C.; Wen, J.; Luo, X.; Huang, C.; Wu, Z.; and Xu, Y. 2023{\natexlab{b}}.
\newblock DICNet: Deep Instance-Level Contrastive Network for Double Incomplete
  Multi-View Multi-Label Classification.
\newblock \emph{arXiv preprint arXiv:2303.08358}.

\bibitem[{Liu et~al.(2015)Liu, Luo, Tao, Xu, and Wen}]{liu2015low}
Liu, M.; Luo, Y.; Tao, D.; Xu, C.; and Wen, Y. 2015.
\newblock Low-rank multi-view learning in matrix completion for multi-label
  image classification.
\newblock In \emph{Proceedings of the AAAI conference on artificial
  intelligence}, volume~29.

\bibitem[{Liu et~al.(2022)Liu, Song, Ma, Ganaa, and Shen}]{liu2022more}
Liu, S.; Song, X.; Ma, Z.; Ganaa, E.~D.; and Shen, X. 2022.
\newblock MoRE: multi-output residual embedding for multi-label classification.
\newblock \emph{Pattern Recognition}, 126: 108584.

\bibitem[{Liu et~al.(2023{\natexlab{c}})Liu, Yuan, Lyu, and
  Feng}]{liu2023label}
Liu, W.; Yuan, J.; Lyu, G.; and Feng, S. 2023{\natexlab{c}}.
\newblock Label driven latent subspace learning for multi-view multi-label
  classification.
\newblock \emph{Applied Intelligence}, 53(4): 3850--3863.

\bibitem[{Liu, Sun, and Feng(2022)}]{liu2022incomplete}
Liu, X.; Sun, L.; and Feng, S. 2022.
\newblock Incomplete multi-view partial multi-label learning.
\newblock \emph{Applied Intelligence}, 52(3): 3289--3302.

\bibitem[{Lyu et~al.(2022)Lyu, Deng, Wu, and Feng}]{lyu2022beyond}
Lyu, G.; Deng, X.; Wu, Y.; and Feng, S. 2022.
\newblock Beyond shared subspace: A view-specific fusion for multi-view
  multi-label learning.
\newblock In \emph{Proceedings of the AAAI Conference on Artificial
  Intelligence}, volume~36, 7647--7654.

\bibitem[{Lyu et~al.(2024{\natexlab{a}})Lyu, Kang, Wang, Li, Yang, and
  Feng}]{lyucommon1}
Lyu, G.; Kang, W.; Wang, H.; Li, Z.; Yang, Z.; and Feng, S. 2024{\natexlab{a}}.
\newblock Common-Individual Semantic Fusion for Multi-View Multi-Label
  Learning.
\newblock In \emph{IJCAI}.

\bibitem[{Lyu et~al.(2024{\natexlab{b}})Lyu, Yang, Deng, and Feng}]{lyu2024vsm}
Lyu, G.; Yang, Z.; Deng, X.; and Feng, S. 2024{\natexlab{b}}.
\newblock L-VSM: Label-Driven View-Specific Fusion for Multiview Multilabel
  Classification.
\newblock \emph{IEEE Transactions on Neural Networks and Learning Systems}.

\bibitem[{Sanghavi and Verma(2022)}]{sanghavi2022multi}
Sanghavi, R.; and Verma, Y. 2022.
\newblock Multi-view multi-label canonical correlation analysis for cross-modal
  matching and retrieval.
\newblock In \emph{Proceedings of the IEEE/CVF Conference on Computer Vision
  and Pattern Recognition}, 4701--4710.

\bibitem[{Sun(2013)}]{sun2013survey}
Sun, S. 2013.
\newblock A survey of multi-view machine learning.
\newblock \emph{Neural computing and applications}, 23: 2031--2038.

\bibitem[{Tan et~al.(2019)Tan, Yu, Wang, Domeniconi, and
  Zhang}]{tan2019individuality}
Tan, Q.; Yu, G.; Wang, J.; Domeniconi, C.; and Zhang, X. 2019.
\newblock Individuality-and commonality-based multiview multilabel learning.
\newblock \emph{IEEE transactions on cybernetics}, 51(3): 1716--1727.

\bibitem[{Wang et~al.(2023)Wang, Wang, Zhang, Jia, Qin, Zuo, Zhang, and
  Dong}]{wang2023segmentalized}
Wang, Z.; Wang, N.; Zhang, H.; Jia, L.; Qin, Y.; Zuo, Y.; Zhang, Y.; and Dong,
  H. 2023.
\newblock Segmentalized mRMR features and cost-sensitive ELM with fixed inputs
  for fault diagnosis of high-speed railway turnouts.
\newblock \emph{IEEE Transactions on Intelligent Transportation Systems}.

\bibitem[{Wu et~al.(2019)Wu, Chen, Hu, Wang, Chang, Wang, and
  Zhang}]{wu2019multi}
Wu, X.; Chen, Q.-G.; Hu, Y.; Wang, D.; Chang, X.; Wang, X.; and Zhang, M.-L.
  2019.
\newblock Multi-View Multi-Label Learning with View-Specific Information
  Extraction.
\newblock In \emph{IJCAI}, 3884--3890.

\bibitem[{Xu, Tao, and Xu(2013)}]{xu2013survey}
Xu, C.; Tao, D.; and Xu, C. 2013.
\newblock A survey on multi-view learning.
\newblock \emph{arXiv preprint arXiv:1304.5634}.

\bibitem[{Xu, Liu, and Geng(2019)}]{xu2019label}
Xu, N.; Liu, Y.-P.; and Geng, X. 2019.
\newblock Label enhancement for label distribution learning.
\newblock \emph{IEEE Transactions on Knowledge and Data Engineering}, 33(4):
  1632--1643.

\bibitem[{Yagi(2011)}]{yagi2011color}
Yagi, Y. 2011.
\newblock Color standardization and optimization in whole slide imaging.
\newblock In \emph{Diagnostic pathology}, volume~6, 1--12. Springer.

\bibitem[{Yelipe, Porika, and Golla(2018)}]{yelipe2018efficient}
Yelipe, U.; Porika, S.; and Golla, M. 2018.
\newblock An efficient approach for imputation and classification of medical
  data values using class-based clustering of medical records.
\newblock \emph{Computers \& Electrical Engineering}, 66: 487--504.

\bibitem[{Yin and Zhang(2023)}]{yin2023multi}
Yin, J.; and Zhang, W. 2023.
\newblock Multi-view multi-label learning with double orders manifold
  preserving.
\newblock \emph{Applied Intelligence}, 53(12): 14703--14716.

\bibitem[{Yuan and Lin(2006)}]{yuan2006model}
Yuan, M.; and Lin, Y. 2006.
\newblock Model selection and estimation in regression with grouped variables.
\newblock \emph{Journal of the Royal Statistical Society Series B: Statistical
  Methodology}, 68(1): 49--67.

\bibitem[{Zhang et~al.(2024)Zhang, Fang, Liang, Zhang, Zhou, Wu, Jiang, and
  Sheng}]{zhang2024efficient}
Zhang, C.; Fang, Y.; Liang, X.; Zhang, H.; Zhou, P.; Wu, J., Xingyu~andYang;
  Jiang, B.; and Sheng, W. 2024.
\newblock Efficient Multi-view Unsupervised Feature Selection with Adaptive
  Structure Learning and Inference.
\newblock In \emph{Proceedings of the 33rd International Joint Conference on
  Artificial Intelligence}, 5443--5452.

\bibitem[{Zhang et~al.(2018)Zhang, Yu, Hu, Zhu, Liu, and
  Wang}]{zhang2018latent}
Zhang, C.; Yu, Z.; Hu, Q.; Zhu, P.; Liu, X.; and Wang, X. 2018.
\newblock Latent semantic aware multi-view multi-label classification.
\newblock In \emph{Proceedings of the AAAI Conference on Artificial
  Intelligence}, volume~32.

\bibitem[{Zhang et~al.(2020{\natexlab{a}})Zhang, Lin, Jiang, Li, Tang, and
  Tan}]{zhangjia2020multi}
Zhang, J.; Lin, Y.; Jiang, M.; Li, S.; Tang, Y.; and Tan, K.~C.
  2020{\natexlab{a}}.
\newblock Multi-label Feature Selection via Global Relevance and Redundancy
  Optimization.
\newblock In \emph{IJCAI}, 2512--2518.

\bibitem[{Zhang and Zhou(2013)}]{zhang2013review}
Zhang, M.-L.; and Zhou, Z.-H. 2013.
\newblock A review on multi-label learning algorithms.
\newblock \emph{IEEE transactions on knowledge and data engineering}, 26(8):
  1819--1837.

\bibitem[{Zhang, Xu, and Zhou(2024)}]{zhang2024realnet}
Zhang, X.; Xu, M.; and Zhou, X. 2024.
\newblock RealNet: A feature selection network with realistic synthetic anomaly
  for anomaly detection.
\newblock In \emph{Proceedings of the IEEE/CVF Conference on Computer Vision
  and Pattern Recognition}, 16699--16708.

\bibitem[{Zhang et~al.(2020{\natexlab{b}})Zhang, Wu, Cai, and
  Philip}]{zhang2020multi}
Zhang, Y.; Wu, J.; Cai, Z.; and Philip, S.~Y. 2020{\natexlab{b}}.
\newblock Multi-view multi-label learning with sparse feature selection for
  image annotation.
\newblock \emph{IEEE Transactions on Multimedia}, 22(11): 2844--2857.

\bibitem[{Zhao et~al.(2022)Zhao, Gao, Lu, and Sun}]{zhao2022learning}
Zhao, D.; Gao, Q.; Lu, Y.; and Sun, D. 2022.
\newblock Learning view-specific labels and label-feature dependence
  maximization for multi-view multi-label classification.
\newblock \emph{Applied Soft Computing}, 124: 109071.

\bibitem[{Zhong, Lyu, and Yang(2024)}]{zhong2024align}
Zhong, Q.; Lyu, G.; and Yang, Z. 2024.
\newblock Align While Fusion: A Generalized Nonaligned Multiview Multilabel
  Classification Method.
\newblock \emph{IEEE Transactions on Neural Networks and Learning Systems}.

\bibitem[{Zhu et~al.(2020)Zhu, Miao, Wang, Zhou, Wei, and
  Zhang}]{zhu2020global}
Zhu, C.; Miao, D.; Wang, Z.; Zhou, R.; Wei, L.; and Zhang, X. 2020.
\newblock Global and local multi-view multi-label learning.
\newblock \emph{Neurocomputing}, 371: 67--77.

\bibitem[{Zhu et~al.(2018)Zhu, Hu, Hu, Zhang, and Feng}]{zhu2018multi}
Zhu, P.; Hu, Q.; Hu, Q.; Zhang, C.; and Feng, Z. 2018.
\newblock Multi-view label embedding.
\newblock \emph{Pattern recognition}, 84: 126--135.

\bibitem[{Zhu(2005)}]{zhu2005semi}
Zhu, X. 2005.
\newblock \emph{Semi-supervised learning with graphs}.
\newblock Carnegie Mellon University.

\bibitem[{Zhu, Li, and Zhang(2015)}]{zhu2015block}
Zhu, X.; Li, X.; and Zhang, S. 2015.
\newblock Block-row sparse multiview multilabel learning for image
  classification.
\newblock \emph{IEEE transactions on cybernetics}, 46(2): 450--461.

\end{thebibliography}

\end{document}